\pdfoutput=1

\documentclass[11pt]{article}

\usepackage{ACL2023}

\usepackage{times}
\usepackage{latexsym}
\usepackage{subcaption} 
\usepackage{float}  
\usepackage{cuted} 
\usepackage[T1]{fontenc}

\usepackage[utf8]{inputenc}
\usepackage{CJKutf8} 
\usepackage{multirow}
\usepackage{multicol}

\usepackage{microtype}

\usepackage{inconsolata}
\usepackage{pgfplots} 
\usepackage{graphicx}
\usepackage{cellspace}
\usepackage{subcaption}
\title{Is Translation Helpful? An Empirical Analysis of Cross-Lingual Transfer in Low-Resource Dialog Generation}

\author{Lei Shen, Shuai Yu \\
  Donghua University \\
  \texttt{lorashen@126.com} \\\And
  Xiaoyu Shen~\thanks{\hspace{1.5 mm}Work done outside Amazon} \\
  Amazon Alexa AI \\
  \texttt{gyouu@amazon.com} \\}

\begin{document}
\maketitle
\begin{abstract}
Cross-lingual transfer is important for developing high-quality chatbots in multiple languages due to the strongly imbalanced distribution of language resources. A typical approach is to leverage off-the-shelf machine translation (MT) systems to utilize either the training corpus or developed models from high-resource languages. In this work, we investigate whether it is helpful to utilize MT at all in this task. To do so, we simulate a low-resource scenario assuming access to limited Chinese dialog data in the movie domain and large amounts of English dialog data from multiple domains. Experiments show that leveraging English dialog corpora can indeed improve the naturalness, relevance and cross-domain transferability in Chinese. However, directly using English dialog corpora in its original form, surprisingly, is better than using its translated version. As the topics and wording habits in daily conversations are strongly culture-dependent, MT can reinforce the bias from high-resource languages, yielding unnatural generations in the target language. Considering the cost of translating large amounts of text and the strong effects of the translation quality, we suggest future research should rather focus on utilizing the original English data for cross-lingual transfer in dialog generation. 
We perform extensive human evaluations and ablation studies. The analysis results, together with the collected dataset, are presented to draw attention towards this area and benefit future research~\footnote{\url{https://github.com/lorashen/cross_lingual_transfer_dialog_generation}}.
\end{abstract}

\section{Introduction}
Dialog systems (chatbots) have made great progress and have achieved close-to-human performances in many scenarios~\cite{su2020moviechats,adiwardana2020towards,shuster2022blenderbot,thoppilan2022lamda,liu2023summary}. However, current state-of-the-art approaches rely on huge amounts of training data, which is only available in English 
and a few high-resource languages--such as Chinese~\cite{zhang-etal-2020-dialogpt}, Japanese~\cite{sugiyama2021empirical} and German~\cite{stefan_schweter_2020_4275046}.

\begin{figure}
    \centering
     \includegraphics[scale=0.8]{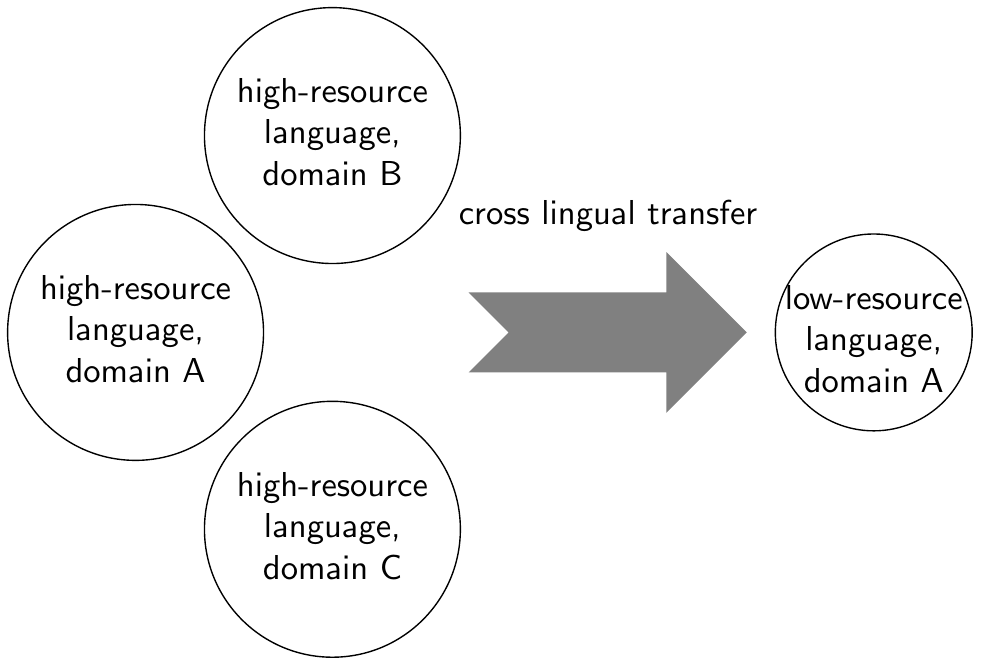}

    \caption{\small Scenario that requires cross-lingual transfer for dialog generation: There is large amounts of dialog data from various domains in a high-resource langauge, but only limited dialog data from one domain in a low-resource language.}
    \label{fig:intro}
\end{figure}

Typically each language develops its own chatbot individually without cross-lingual resource sharing. Repeating this process for all languages is infeasible, as most low-resource languages do not have enough conversational data to support this type of training~\cite{zhao2020low,shen2022low}. Even for high-resource languages, collecting sufficient amount of high-quality data to cover various domains is still costly~\cite{xu2020data,chang2021selectgen}. Therefore, we believe cross-lingual transfer is crucial for efficiently developing chatbots in multiple languages, through which the same resource can be reused across languages. Figure~\ref{fig:intro} illustrates the scenario that we are targeting at in this paper. This is a common scenario for most low-resource languages since usually we can only afford collecting high-quality dialogs for one specific domain.

There have been many studies on cross-lingual transfer for classification tasks~\citep{Hu-2020,Jiang-2020,Ruder-2021,Ding-2021}. For generation tasks, however, much less attention has been paid to it and the results are far from satisfactory~\cite{cao2020jointly,chang2020unsupervised,Chen-2021,vzagar2021cross,shen2023xpqa}. The challenge is especially prominent in dialog generation, as different language users have different habits of conversing. For example, the typical conversation ``-How are you? -Fine, and you?” in English can be very unnatural when translated into other languages such as Chinese or Japanese, because their speakers do not usually greet each other in this way~\cite{zhang2021dataset,zhang2022mdia}. This is usually not a big problem for understanding tasks but crucial for dialog generation if we would like to produce human-like, culturally grounded conversations.

In this work, we investigate the
performance of several baseline methods for cross-lingual transfer in dialog generation. To simulate a low-resource scenario, we collect limited Chinese conversational data related to the movie domain and large amounts of English conversational data related to various domains as our training data. The test data cover three additional domains--music, books and technology--so that we can test the domain transferability of developed models~\footnote{Even if Chinese is not a low-resource language itself, we choose Chinese as our target language because (1) it has large available corpus from various domains to easily build this setup, and (2) it comes from a different language family with a different writing system from English, so as to mimic the realistic scenario of most low-resource languages.}. We construct this benchmark dataset in order to see \emph{how can we effectively leverage the English data to benefit us in developing a good Chinese chatbot}.

We compare three types of baseline cross-lingual transfer techniques: (1) \emph{translate-train}, which translates the English training data into Chinese first and finetunes a Chinese-centric chatbot on it; (2) \emph{translate-test}, which trains an English-centric chatbot first and uses a translator at inference time; and (3) \emph{multilingual finetune}, which simply finetunes on English data followed by Chinese data regardless of their vocabulary difference. \emph{Multilingual finetune} has been a common practice in classification tasks but rarely applied for generation tasks~\cite{alabi-etal-2020-massive,Ruder-2021}. We find that \emph{translate-train} consistently outperforms \emph{translate-test} but both suffer from the translationese problem. \emph{Multilingual finetune}, surprisingly, perform the best with as few as 500 Chinese dialogs available for training. The advantage further grows with increasing Chinese dialogs. 



Our contribution can be summarized as follow:
(1) We construct a benchmark dataset covering various domains for studying cross-lingual transfer in dialog generation, which can be used for further studies.
(2) We compare baseline models through comprehensive human evaluations for both in-domain and out-of-domain performances.
(3) We conduct extensive experiments to study the effects of various factors such as the translation quality and the training set size. Results and analysis are shared to benefit future research.

\section{Data Collection}
We collect a benchmark to simulate the scenario in Figure~\ref{fig:intro}.
As mentioned, we choose English as the source high-resource language and Chinese as the target low-resource language.

\noindent\textbf{Source.} We collect English dialogs from Reddit\footnote{https://www.reddit.com/} and Chinese ones from Douban\footnote{https://www.douban.com/}, both being popular social forums in the US and China respectively.

\noindent\textbf{Domain.} We choose four domains that are shared between Reddit and Douban: movies, music, books, and technology. The English dialogs are collected from these four subreddits, and Chinese ones are from the corresponding Douban groups. To simulate the scenario where the English corpora is large enough to provide various domains whereas the Chinese corpora has only limited data in one domain, we collect equal number of dialogs from each domain for English. For Chinese, we collect the training set only from the \emph{movie domain}, and the test set from all the four domains.

\noindent\textbf{Preprocessing.} We filter the sentences that fulfills any of the following conditions: (1) too short (less than 5 words for English and 6 characters for Chinese); (2) too long (more than 128 words); (3) contains URLs or offensive words identified
by phrase matching against a large blocklist; (4) from a known bot; (5) the response contains words that repeat over 3 times.

\noindent\textbf{Size.} In our base setting, we use 400k/20k English dialogs and 500/50 Chinese dialogs for training/validation. The test set contains 500 Chinese dialogs from each of the 4 domains.


\begin{table*}[!tb]
\small
\centering
\renewcommand{\arraystretch}{0.9}
	\setlength\cellspacetoplimit{3pt}
\setlength\cellspacebottomlimit{0pt}
	\begin{tabular}{Sc|Sc Sc Sc|Sc Sc Sc Sc}
\hline
  \multirow{2}{*}{} & \multicolumn{3}{c|}{Automatic Evaluation} & \multicolumn{4}{c}{Human Evaluation} \\
\cline{2-8}
 \multirow{2}{*}{} & BLEU2  &Distinct-1 & Distinct-2 &naturalness  &diversity &relevance &Overall\\
\hline
FT & 4.12  & \underline{0.939}& 0.895& 2.67 & 2.69 & 2.24 & 2.24\\
\cline{1-8}
Train\underline{\hspace{0.5em}}Zero & 5.33 & 0.887&\textbf{0.946} & 2.56 & 2.91 & 2.58 &2.34\\
Train\underline{\hspace{0.5em}}Few & \underline{5.37}  &0.931&0.923&2.76 & 2.86 & 2.73 & 2.69 \\
\cline{1-8}
Test\underline{\hspace{0.5em}}Zero & 4.90  &0.876 &\underline{0.935}& 2.28 & \underline{2.98} & 2.30&2.26 \\
Test\underline{\hspace{0.5em}}Few & 
 4.58  &0.906& 0.922& 2.45 & \textbf{3.00} & 2.11 &2.17 \\
\cline{1-8}
 Multi-FT & \underline{5.37} &0.936&0.926& \underline{3.06} & 2.93 & \textbf{2.78}  &\underline{2.95}\\
\cline{1-8}
 GPT-Chinese & \textbf{5.49} & \textbf{0.974}&0.918&\textbf{3.40} & 2.94 & \underline{2.76} &\textbf{3.07} \\

\hline
\end{tabular}
\caption{\small The results for seven methods. For this table, we fix the training set size as 400k for English, and 500 for Chinese. The best score is in bold, and the one with underline is the second best.}
\label{tab:result}
\end{table*}

\section{Approaches}
We implement three popular types of methods for cross-lingual transfer: (1) translate-train, (2) translate-test and (3) multilingual-finetune. The first two are further tested in the zero-shot setting without Chinese dialogs, and the few-shot setting with limitd Chinese dialogs for finetuning. 

\noindent\textbf{Translate-Train}
The translate-train approach first translates the English training corpora into Chinese to train a Chinese-centric chatbot. In the zero-shot setting (Train\_Zero), the model is only trained on the translated corpora. In the few-shot setting (Train\_Few), the model is trained on the translated corpora then finetuned on the Chinese corpora. 

\noindent\textbf{Translate-Test}
The translate-test approach trains an English-centric chatbot. During inference time, we translate the Chinese context into English, generate its response, then translate the response back into Chinese.  In the zero-shot setting (Test\_Zero), the model is only trained on the original English corpora. In the few-shot setting (Test\_Few), the model is trained on the original corpora followed by the translated Chinese corpora.




\noindent\textbf{Multilingual-finetune}
The multilingual-finetune (multi-FT) approach trains the model on the original English corpora then finetunes on the Chinese corpora regardless of without leveraging any external translators. This approach only applies to the few-shot setting. In the zero-shot setting, the model will only generate English responses as it is trained only on English responses.

We further compare with two more methods: (1) Chinese-only finetune (FT), which only finetunes on the Chinese dialog corpora without cross-lingual transfer and (2) GPT-Chinese, which finetunes a pretrained Chinese GPT-2 chatbot\footnote{https://github.com/yangjianxin1/GPT2-chitchat} using the Chinese corpora. This can serve as an upper bound of cross-lingual performance since it accessed large amounts of dialogs in the target language.

\section{Experiments}

\subsection{Settings}
We initialize all approaches with the pretrained MT5-base model, which is a multilingual model that support 101 languages~\cite{xue-etal-2021-mt5}, to keep the comparison fair.
We use MarianMT as the basic translation method~\cite{mariannmt}, as it is a widely used machine translation tool that provides more than 1000 translation models. Hyperparameter details are in the appendix.


\subsection{Evaluation Metric}
We employ both automatic and human evaluations to assess the performance of the compared methods.  We use BLEU, Distinct-1 and Distinct-2 as the automatic evaluation metrics.

\textbf{BLEU} measures the n-gram overlap between predicted response and the target response~\cite{papineni2002bleu}. We report the bigram BLEU-2 score using the sacreBLEU toolkit\footnote{https://github.com/mjpost/sacrebleu}.

\textbf{Distinct-1/2} measure the generation diversity, i.e., the percentage of distinct uni- or bi-grams in generated words~\cite{li-etal-2016-diversity,shen2018nexus}.

\begin{figure*}[!htb]

\centering
 \begin{subfigure}{0.32\textwidth}

		\centering
	\begin{tikzpicture}[scale=0.59] 
	
	\begin{axis}[
	ylabel=BLEU2, 
	tick align=outside, 
	ymin=3.2,
	ytick={3.5,4,4.12,4.5,5,5.37},
	yticklabels={3.5,4,FT,4.5, 5,Multi-FT},
	xtick=data,
	xticklabels={10,50,100, 200,MarianMT},
	legend style={at={(0.8,0.35)},anchor=north} 
	]
	
	\addplot[sharp plot,mark=triangle*,ultra thick,green] plot coordinates { 
		(10,4.76)
		(50,4.77)
		(100,4.83)
		(150,5.03)
		(200, 5.37)
	};
	
	\addlegendentry{Train\underline{\hspace{0.5em}}Few}
	
	\addplot[sharp plot,mark=x,blue,ultra thick, dashed] plot coordinates { 
		(10,4.52)
		(50,4.69)
		(100,4.75)
		(150,4.80)
		(200, 5.33)
	};
	
	\addlegendentry{Train\underline{\hspace{0.5em}}Zero}

 \addplot[sharp plot,mark=square*,purple, ultra thick,dash dot] plot coordinates { 
		(10,4.14)
		(50,4.36)
		(100,4.41)
		(150,4.51)
		(200, 4.90)
	};
 \addlegendentry{Test\underline{\hspace{0.5em}}Zero}
	\definecolor{portlandorange}{rgb}{1.0, 0.35, 0.21}
	\addplot[sharp plot,mark=*,portlandorange, ultra thick,densely dotted] plot coordinates { 
		(10,4.07)
		(50,4.18)
		(100,4.25)
		(150,4.31)
		(200, 4.58)
	};
	
	\addlegendentry{Test\underline{\hspace{0.5em}}Few}
	\addplot[mark=none, black,ultra thick,dash dot] coordinates {(0,5.37) (200,5.37)};
	\addplot[mark=none, gray,ultra thick,dashed] coordinates {(0,4.12) (200,4.12)};
	
	\end{axis}
	\end{tikzpicture}
 \caption{Translation Qualities.}
\label{fig:trans}
\end{subfigure}
\begin{subfigure}{0.32\textwidth}
\centering
 \begin{tikzpicture}[scale=0.59] 
\begin{axis}[
ylabel=BLEU2, 
tick align=outside, 
ymin=3.2,
ytick={3.5,4,4.5,5,5.5},
yticklabels={3.5,4,4.5, 5,5.5},
xtick=data,
xticklabels={4k,40k,400k,1m},
legend style={at={(0.8,0.43)},anchor=north} 
]

\addplot[sharp plot,mark=x,orange,ultra thick,dash pattern={on 7pt off 2pt on 1pt off 3pt}] plot coordinates { 
	(10,4.58)
	(50,4.97)
	(100,5.33)
	(150,5.39)
};

\addlegendentry{Train\underline{\hspace{0.5em}}Zero}

\addplot[sharp plot,mark=*,blue,ultra thick,dash dot] plot coordinates { 
	(10,4.98)
	(50,5.10)
	(100,5.37)
	(150,5.46)
};
\addlegendentry{Train\underline{\hspace{0.5em}}Few}

\addplot[sharp plot,mark=o,purple,ultra thick,dashed] plot coordinates { 
	(10,4.2)
	(50,4.8)
	(100,4.9)
	(150,5.17)
};

\addlegendentry{Test\underline{\hspace{0.5em}}Zero}

\addplot[sharp plot,mark=triangle*,cyan,ultra thick,densely dotted] plot coordinates { 
	(10,3.46)
	(50,4.27)
	(100,4.58)
	(150,4.63)
};

\addlegendentry{Test\underline{\hspace{0.5em}}Few}

\addplot[sharp plot,mark=square*,green,ultra thick] plot coordinates { 
	(10,4.87)
	(50,5.14)
	(100,5.37)
	(150,5.51)
};

\addlegendentry{Multi-FT}
\end{axis}
\end{tikzpicture}
	\caption{English Training Sizes.}
 \label{fig:sizea}
\end{subfigure}
\begin{subfigure}{0.32\textwidth}
\centering
\begin{tikzpicture}[scale=0.59] 

\begin{axis}[
ylabel=BLEU2, 
tick align=outside, 
ymin=3.2,
ytick={3,3.5,4,4.5,5,5.5,6},
yticklabels={3,3.5,4,4.5, 5,5.5,6},
xtick=data,
xticklabels={5h,5k,1w},
legend style={at={(0.8,0.35)},anchor=north} 
]
\addplot[sharp plot,mark=x,orange,ultra thick] plot coordinates { 
	(10,4.12)
	(50,5.12)
	(100,5.5)
};

\addlegendentry{FT}

\addplot[sharp plot,mark=*,blue,ultra thick,densely dotted] plot coordinates { 
	(10,5.37)
	(50,5.49)
	(100,5.8)
};

\addlegendentry{Train\underline{\hspace{0.5em}}Few}

\addplot[sharp plot,mark=triangle*,purple,ultra thick,dash dot] plot coordinates { 
	(10,4.58)
	(50,4.45)
	(100,4.34)
};

\addlegendentry{Test\underline{\hspace{0.5em}}Few}
\addplot[sharp plot,mark=square*,green,ultra thick,dash pattern={on 7pt off 2pt on 1pt off 3pt}] plot coordinates { 
	(10,5.37)
	(50,5.69)
	(100,5.99)
};

\addlegendentry{Multi-FT}

\end{axis}
\end{tikzpicture}
	\caption{Chinese Training sizes.}
 \label{fig:size}
\end{subfigure}
\caption{\small BLEU-2 results by varying the translation qualities of translator models, as well as the training sizes.}
\end{figure*}
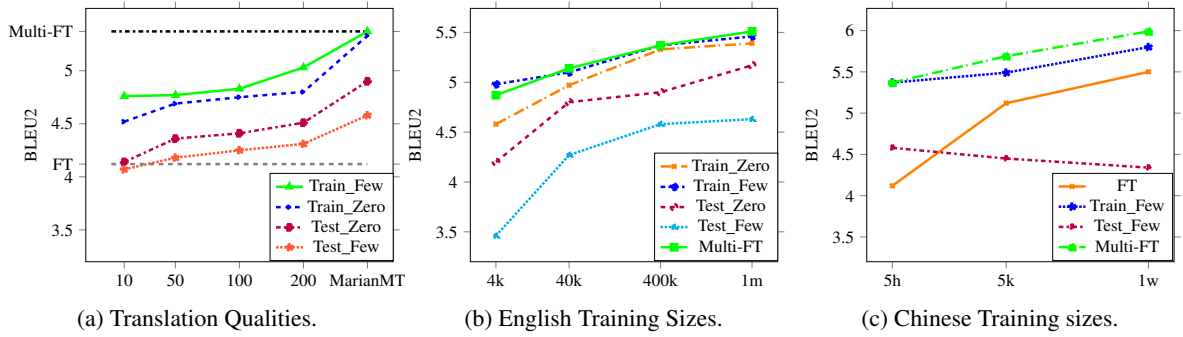
For human evaluation, we randomly select 200 dialogue contexts from the testset and generate responses using the compared methods. The annotators are
asked to rate, using a score of 1 to 5, the response quality from four perspectives--\textbf{Naturalness}, \textbf{Diversity}, \textbf{Coherence}, and \textbf{Overall}. A higher score indicates better quality. 
\subsection{Analysis}
\paragraph{Overall Result}

The results of the seven approaches evaluated in the movie domain are shown in Table~\ref{tab:result}. According to the human evaluation, 
GPT-Chinese performs the best and excels especially on the naturalness score. This is expected since it is pretrained on 500k Chinese dialogs and has learnt more about how to produce more natural responses. The diversity of the method \emph{FT} is worse than the others, which suggests finetuning only on a small Chinese corpora may fail to generate more diverse responses.

The \emph{translate-test} methods perform worse than the \emph{translate-train} methods, because the reliance on an external translator in the inference time might reinforce the error propagation. In most metrics, \emph{translate-test} methods are even worse than the \emph{FT} baseline, suggesting they might not be a good way to consider for cross-lingual transfer.

\emph{Multi-FT} outperforms all other cross-lingual transfer approaches, especially on the naturalness score. This is interesting since its first-stage training does not update any Chinese word embeddings but only the upper-level encoder-decoders. Only in the second-stage finetuning, the Chinese word embeddings get updated to adapt to upper-level encoder-decoders. This suggests the upper-level parameters might be more important and can learn universal conversational knowledge beyond for one fixed language. Further finetuning on a small target-lanugage corpus (500 dialogs) is enough to adapt to new vocabularies. Similar findings have been found for classification tasks~\cite{alabi-etal-2020-massive}.

\definecolor{tuftsblue}{rgb}{0.28, 0.57, 0.81}
\definecolor{lightgreen}{rgb}{0.56, 0.93, 0.56}
\definecolor{brickred}{rgb}{0.8, 0.25, 0.33}
\definecolor{lavenderpurple}{rgb}{0.59, 0.48, 0.71}
\definecolor{princetonorange}{rgb}{1.0, 0.56, 0.0}
\definecolor{stildegrainyellow}{rgb}{0.98, 0.85, 0.37}	
\definecolor{lightseagreen}{rgb}{0.13, 0.7, 0.67}
\definecolor{darkpastelred}{rgb}{0.76, 0.23, 0.13}
\definecolor{palered-violet}{rgb}{0.86, 0.44, 0.58}
\definecolor{lightblue}{rgb}{0.68, 0.85, 0.9}
\definecolor{lightskyblue}{rgb}{0.53, 0.81, 0.98}
\pgfplotstableread[row sep=\\,col sep=&]{
    interval & carT  & carD    & carR    &  busT    & busD    & busR    &  walkT         \\
    movie &  2.24 &  2.34 &  2.68 &  2.26 & 2.17 & 2.95 &  3.07  \\
    music &  2.06 &  2.32 &  2.58 &  2.23 & 2.14 & 2.79 &  2.88  \\
    books & 2.09 & 2.35 & 2.64 &  2.20 & 2.16 & 
    2.85 &  2.96  \\
    technology & 2.04 & 2.33 & 2.60 &  2.18 & 2.12 & 
    2.81 &  2.90  \\
    }\mydata
\pgfplotstableread[row sep=\\,col sep=&]{
	interval & carT  & carD    & carR    &  busT    & busD    & busR    &  walkT         \\
	movie  &  2.24 &  2.34 &  2.68 &  2.26 & 2.17 & 2.95 &  3.07  \\
	music  &  2.24 &  2.34 &  2.68 &  2.26 & 2.17 & 2.95 &  3.07  \\
	books  &  2.24 &  2.34 &  2.68 &  2.26 & 2.17 & 2.95 &  3.07  \\
	technology  &  2.24 &  2.34 &  2.68 &  2.26 & 2.17 & 2.95 &  3.07  \\
}\mydataBg

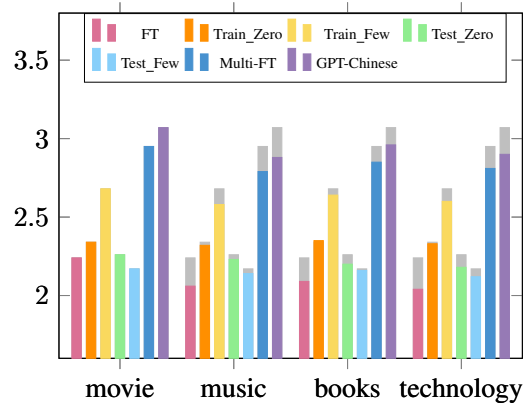
\begin{figure}[ht]
\centering
  \begin{tikzpicture}[scale = 0.95]
            \begin{axis}[
                    ybar,
                    bar width=.13cm,
                    width=.5\textwidth,
                    height=.4\textwidth,
                    legend columns = 4,
                    legend style={at={(0.51,1)},
                        anchor=north, font=\tiny},
                    symbolic x coords={movie,music,books,technology},
                    xtick=data,
                    enlarge x limits={abs=6.5*\pgfplotbarwidth},
                    ymin=1.6,ymax=3.8,
                ]
                   \addplot[lightgray,fill] table[x=interval,y=carT]{\mydataBg};
                \addplot[lightgray,fill] table[x=interval,y=carD]{\mydataBg};
                \addplot[lightgray,fill] table[x=interval,y=carR]{\mydataBg};
                \addplot[lightgray,fill] table[x=interval,y=busT]{\mydataBg};
                \addplot[lightgray,fill] table[x=interval,y=busD]{\mydataBg};
                \addplot[lightgray,fill] table[x=interval,y=busR]{\mydataBg};
                \addplot[lightgray,fill] table[x=interval,y=walkT]{\mydataBg};
              \end{axis}
            \begin{axis}[
                    ybar,
                    bar width=.13cm,
                    width=.5\textwidth,
                    height=.4\textwidth,
                    legend columns = 4,
                    legend style={at={(0.51,1)},
                        anchor=north, font=\tiny},
                    symbolic x coords={movie,music,books,technology},
                    xtick=data,
                    enlarge x limits={abs=6.5*\pgfplotbarwidth},
                    ymin=1.6,ymax=3.8,
                ]

                \addplot[palered-violet,fill] table[x=interval,y=carT]{\mydata};
                
                \addplot[princetonorange,fill] table[x=interval,y=carD]{\mydata};
                \addplot[stildegrainyellow,fill] table[x=interval,y=carR]{\mydata};
                \addplot[lightgreen,fill] table[x=interval,y=busT]{\mydata};
                \addplot[lightskyblue,fill] table[x=interval,y=busD]{\mydata};
                \addplot[tuftsblue,fill] table[x=interval,y=busR]{\mydata};
                \addplot[lavenderpurple,fill] table[x=interval,y=walkT]{\mydata};
               
                \legend{FT, Train\underline{\hspace{0.5em}}Zero, Train\underline{\hspace{0.5em}}Few, Test\underline{\hspace{0.5em}}Zero, Test\underline{\hspace{0.5em}}Few, Multi-FT, 
                GPT-Chinese}
            \end{axis}
    \end{tikzpicture}
\caption{\small Overall Human Scores for Four Domains. The grey color indicates the drop compared with the movie domain.}
\label{fig:domain}
\end{figure}

\paragraph{Cross-Domain}
Figure~\ref{fig:domain} shows the model performances in the other three domains. In general all models drop in the other domains. \emph{FT} drops especially more, which indicates cross-lingual transfer could improve the domain transferability of the model. \emph{Multi-FT} still performs the best among cross-lingual approaches, but it drops more than translation-based methods.

\paragraph{Translation Quality}
To simulate translators with different qualities, we collect different sizes of English/Chinese data from WMT17\footnote{https://www.statmt.org/wmt17/} to train different translation models (all models are nitialized from MT5-base). 
Fig~\ref{fig:trans} shows the BLEU-2 scores with different translation qualities. When we replace MarianMT with other translation models that are not well trained, the BLEU2 scores for all the translation-based methods decrease by a large margin. We conclude that translation quality influences the performance of the methods considerably. If the quality of the translator is bad, the model can even underperform the \emph{FT} baseline without any cross-lingual transfer. Considering that most low-resource languages do not have high-quality MT systems yet~\cite{adelani2022few}, this further implies we should rather focus on translation-free approaches for this task.

\paragraph{Training Set Size}
Fig~\ref{fig:sizea} and~\ref{fig:size} further shows the BLEU-2 scores with varying sizes of English and Chinese training data. As the training size of English corpora becomes larger, all cross-lingual-transfer methods perform consistently better. Zeroshot approaches are affected more than fewshot performances as they rely solely on the English corpora to train. When increasing the Chinese corpus, all models perform better except \emph{Test-Few}. This is possible because \emph{Test-Few} is trained on the translated Chinese corpus which is not guaranteed to be cycle-consistent when translated back. Therefore, its training objective does not fully align with our inference-time objective and increasing the Chinese corpura size might not help.  The advantage of \emph{Multi-FT} over the other translation-based methods also improves with more Chinese data. Considering further the cost of translating the corpus, \emph{Multi-FT} seems to be a better baseline for cross-lingual transfer than translation-based methods.


\subsection{Conclusion}
In this work, we construct a benchmek to systematically study the task of cross-lingual transder for dialog generation. We conduct extensive experiments and ablation studies to understand the performance of popular baseline methods. The results suggest that directly training on high-resource-language data then finetuning on low-resource-language data yield a very strong baseline, improving both the naturelness, relevance and domain transferability. An external translator might not be necessary.

\section*{Limitations}
As we concluded, by training on the original English corpora, the naturalness and relevance of the generated Chinese responses can be improved. However, when training models on English corpora, the Chinese embeddings are not updated, and only encoder/decoder layers are updated. Thus, the Chinese embeddings might not be compatible with encoder/decoder layers after training. We plan to investigate how to alleviate this problem during training in the future. Furthermore, we only studied two languages for the 4 considered domains. To which extent the results drawn from this study also apply to other languages and domains is still uncertain.

\bibliography{custom}
\bibliographystyle{acl_natbib}

\appendix

\section{Hyperparameter Details}
The learning rate is 1e-4 for large English training set and 1e-5 for small Chinese training set. The maximum sequence length of context and response is set to 128. The batch size is 16 for MT5-base models. The training epoch is set to 3 for datasets larger than 400k and 9 for smaller datasets. The ADAM optimizer is used. We use top-k top-p sampling for decoding, because it is often used in real generation scenario in order to improve diversity. Top\underline{\hspace{0.5em}}k is set to 3, and top\underline{\hspace{0.5em}}p to 0.9. For each experiment, we run three times to get the mean scores of the automatic evaluation metrics.

\section{Generated Dialog Samples}

We show some generated dialog samples for the best three methods--Train\underline{\hspace{0.5em}}Few, Multi\underline{\hspace{0.5em}}FT, and GPT-Chinese--in Table~\ref{tab:sample}. As we can see, the response that Train\underline{\hspace{0.5em}}Few generated has a grammatical error "\begin{CJK}{UTF8}{gbsn}那片电影\end{CJK}" (that piece of film), which might be brought from the translation step during training. The response of Multi\underline{\hspace{0.5em}}FT generated is natural and relevant. The response of GPT-Chinese is more fluent for dialog scenario than the other two.
\begin{table}[!t]
\centering
\begin{tabular}{p{3em}p{15em}}
\hline
Context& \begin{CJK}{UTF8}{gbsn}今天去看了，我觉得和 9.3分的评价相去甚远。8.3分的评价更合适点。这片更多的看点还是在特技的炫耀上，除此之外，我真没有太多感动。在对时间空间的阐述上，我反而觉得《彗星来的那一夜》更精彩。\end{CJK}\\
& (I went to watch it today, and I think it is far from the rating of 9.3. 8.3 is more appropriate. This film is more focused on the special effects. Besides of that, I am not touched. I think "Coherence" is more wonderful for the explanation of time and space.)\\
\cline{1-2}
Response& Train\underline{\hspace{0.5em}}Few:\begin{CJK}{UTF8}{gbsn}你觉得\textcolor{red}{那片电影}的评分比其他的电影好?\end{CJK}\\
 & (Do you think \textcolor{red}{that piece of film}'s score is better than other films?) \\
 & Multi\underline{\hspace{0.5em}}FT:\begin{CJK}{UTF8}{gbsn}我觉得这片更精彩,我感觉挺不错的。\end{CJK}\\
 & (I think this film is more wonderful. I think it's very good.)\\
 & GPT-Chinese:\begin{CJK}{UTF8}{gbsn}我也看了，那时候真的很感动\end{CJK}\\
 & (I also watched it. I was really touched at that time.)\\

\hline
\end{tabular}
\caption{Generated dialog samples for method Train\underline{\hspace{0.5em}}Few, Multi\underline{\hspace{0.5em}}FT, and GPT-Chinese. The colour red represents grammatical error.}
\label{tab:sample}
\end{table}

\section{The Human Annotation Instructions}
We depict the sketchy definitions of the four perspectives for human annotation here.

\noindent\textbf{Naturalness} 

Score 1: The response include totally unreadable sentences. 

Score 2: The response is readable, but have some grammatical errors, or translationese problems. 

Score 3: The expression of the response is natural without error. 

Score 4: The response is fluent and seemed made by native speakers.

Score 5: The response is quite natural and fluent.

\noindent\textbf{Diversity}

Score 1: The response include repeated words.

Score 2: It is general response.

Score 3: The response is not general, and has some diversity.

Score 4: There are diverse words in the response.

Score 5: There are a lot of diverse words in the response.

\noindent\textbf{Coherence}

Score 1: The response not related to the context at all.

Score 2: The respones is only a little related to the context, or has conflict with the context.

Score 3: The response is related, and does not have conflict with the context.

Score 4: The response is related to the context and is the continuation of the topic in the context.

Score 5: The response is closely related to the context and is all about the topic in the context.

\noindent\textbf{Overall}

Score 1: The response is not related to context at all, or it is unreadable, or it contains repeated words.

Score 2: The response is not related, or it is unnatural, or it is quite general.

Score 3: The response is related, natural, and not general.

Score 4: The response is related or closely related, and quite natural, and not general. 

Score 5: The response is closely related, very fluent, and diverse.

\label{sec:appendix}

\end{document}